\documentclass[journal]{IEEEtran} 
\usepackage{setspace}
\usepackage{cite}
\usepackage[pdftex]{graphicx} 
\graphicspath{ {./image/} }
\usepackage{amsmath}
\usepackage{dsfont}
\usepackage{array}
\ifCLASSOPTIONcompsoc
\usepackage[caption=false,font=normalsize,labelfont=sf,textfont=sf]{subfig}
\else
\usepackage[caption=false,font=footnotesize]{subfig}
\usepackage{diagbox}
\usepackage{booktabs} 
\usepackage{ wasysym } 
\usepackage{xcolor}
\PassOptionsToPackage{cmyk}{xcolor}

\fi
\allowdisplaybreaks
\usepackage{color}
\usepackage{fixmath}
\usepackage{float}
\usepackage{stfloats}
\usepackage{arydshln} 
\usepackage[]{changes}
\usepackage[hyphens]{url}
\usepackage{textcomp} 
\usepackage{adjustbox} 
\usepackage{textcase}
\usepackage{enumerate}
\usepackage{footnote}
\usepackage{xparse}
\usepackage{amssymb}
\usepackage{amsmath}
\usepackage{amsmath, bm}
\usepackage{amsfonts}
\RequirePackage{xspace}
\RequirePackage{graphics}
\RequirePackage{textcomp}
\usepackage{keyval}
\usepackage{xspace}
\usepackage{mathrsfs}
\usepackage{paralist}
\usepackage{lipsum}
\usepackage{listings}
\usepackage{multirow}
\usepackage{array}
\usepackage{pifont} 
\usepackage{stmaryrd}
\usepackage{textcase}
\usepackage{makecell}
\usepackage{stackengine}%
\usepackage[T1]{fontenc}
\usepackage[utf8]{inputenc}
\usepackage[english]{babel}
\usepackage{amsthm}
\usepackage{accents}
\usepackage{algorithm,algpseudocode}
\newtheorem{theorem}{Theorem}

\usepackage{balance}
\RequirePackage{tikz}
\usetikzlibrary{arrows,automata,shapes,calc,through,decorations.pathmorphing,decorations.fractals,chains,shapes.multipart}
\usepackage{arydshln} 
\usepackage{etoolbox} 
\usepackage{circuitikz}
\usepackage{booktabs}
\usepackage{threeparttable, tablefootnote}

\newcommand{\Abf}{\boldsymbol{A}}

\newcommand{\Bbf}{\boldsymbol{B}}

\newcommand{\Gbf}{\boldsymbol{G}}

\newcommand{\Ibf}{\boldsymbol{I}}

\newcommand{\Ncal}{\mathcal{N}}

\newcommand{\Rbb}{\mathbb{R}}
\newcommand{\Rcal}{\mathcal{R}}

\newcommand{\Sbb}{\mathbb{S}}

\newcommand{\ubf}{\boldsymbol{u}}

\newcommand{\xbf}{\boldsymbol{x}}
\newcommand{\Xbf}{\boldsymbol{X}}

\newcommand{\Ybf}{\boldsymbol{Y}}

\newcommand{\T}{^{\!\top}}
\newcommand{\Tcal}{\mathcal{T}}

\makeatletter
\DeclareRobustCommand{\cev}[1]{%
	\mathpalette\do@cev{#1}%
}
\newcommand{\do@cev}[2]{%
	\fix@cev{#1}{+}%
	\reflectbox{$\m@th#1\vec{\reflectbox{$\fix@cev{#1}{-}\m@th#1#2\fix@cev{#1}{+}$}}$}%
	\fix@cev{#1}{-}%
}
\newcommand{\fix@cev}[2]{%
	\ifx#1\displaystyle
	\mkern#2 1mu
	\else
	\ifx#1\textstyle
	\mkern#2 3mu
	\else
	\ifx#1\scriptstyle
	\mkern#2 2mu
	\else
	\mkern#2 2mu
	\fi
	\fi
	\fi
}

\usepackage{stackengine}
\stackMath

\begin{document}
	
\title{Optimal Multi-Robot Motion Planning\\ via Parabolic Relaxation} 
	
\author{Changrak Choi$^{1}$, Muhammad Adil$^{2}$, Amir Rahmani$^{1}$, and Ramtin Madani$^{2}$
\thanks{
$^{1}$ Changrak Choi and Amir Rahmani are with NASA Jet Propulsion Lab, California Institute of Technology, USA 
{\tt\small \{changrak.choi, amir.rahmani\}@jpl.nasa.gov}.

$^{2}$ Muhammad Adil and Ramtin Madani are with the Department of Electrical Engineering, University of Texas at Arlington, USA {\tt\small \{muhammad.adil, ramtin.madani\}@uta.edu}.

This work is supported by the Jet Propulsion Laboratory, California Institute of Technology, under a Contract with the National Aeronautics and Space Administration (NASA).
}
}

\maketitle

\begin{abstract}
Multi-robot systems offer enhanced capability over their monolithic counterparts, but they come at a cost of increased complexity in coordination. 
To reduce complexity and to make the problem tractable, multi-robot motion planning (MRMP) methods in the literature adopt de-coupled approaches that sacrifice either optimality or dynamic feasibility. 
In this paper, we present a convexification method, namely ``parabolic relaxation'', to generate optimal and dynamically feasible trajectories for MRMP in the coupled joint-space of all robots. 
We leverage upon the proposed relaxation to tackle the problem complexity and to attain computational tractability for planning over one hundred robots in extremely clustered environments. 
We take a multi-stage optimization approach that consists of i) mathematically formulating MRMP as a non-convex optimization, ii) lifting the problem into a higher dimensional space, iii) convexifying the problem through the proposed computationally efficient parabolic relaxation, and iv) penalizing with iterative search to ensure feasibility and recovery of feasible and near-optimal solutions to the original problem. 
Our numerical experiments demonstrate that the proposed approach is capable of generating optimal and dynamically feasible trajectories for challenging motion planning problems with higher success rate than the state-of-the-art, yet remain computationally tractable for over one hundred robots in a highly dense environment.
\end{abstract}
	

\section{Introduction}

Multi-robot systems offer enhanced flexibility and robustness over their monolithic counterparts that make them attractive to a wide range of applications \cite{rossi2018review, chung2018survey}. Surveillance, search exploration, and warehouse automation are examples of areas that benefit from multi-robot systems' ability to adapt, reconfigure, be resilient to failure, and replenish over time. However, these benefits come at a cost of increased complexity in the operation. Multi-robot systems pose challenges such as task assignment, motion coordination, data communication, and world model synchronization that need to be handled.



This work addresses the challenges involved with multi-robot motion planning (MRMP). MRMP is the problem of how best to move a team of robots from their current states to given target states while satisfying constraints such as collision avoidance. It is a natural extension of the single-robot motion planning problem which itself has PSPACE-complete computational complexity \cite{canny1988some}. The problem becomes significantly harder when scaled from a single to multi-robot. The computational complexity of MRMP is PSPACE-hard even for a simple robot geometry \cite{solovey2016hardness}, and it becomes NEXPTIME when dynamics of the robot is considered \cite{johnson2016novel, johnson2018relationship}.


\begin{figure}[t!]
 \centering
 \includegraphics[width=1.0\columnwidth]{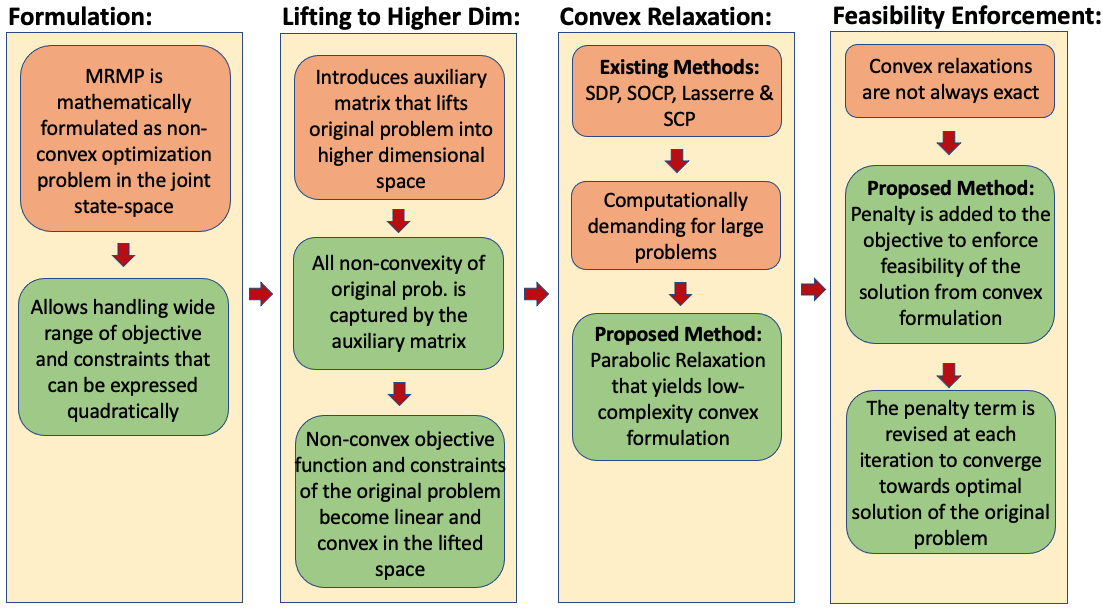}
 \caption{We propose multi-stage optimization approach for solving multi-robot motion planning problem (MRMP) in the joint-space via Parabolic Relaxation. Our approach ensure that the method generates both dynamically feasible and optimal solution to the MRMP problem in the joint-space of all robots, while remaining computationally tractable.
 }
 \label{fig:run_time}
\end{figure}

Current state-of-the-art approaches to MRMP attain computational tractability by making simplifications that sacrifice different aspects of the solution quality. Local reactive methods is highly scalable and fast but far from optimal and susceptible to deadlock \cite{guy2009clearpath, lalish2012distributed, van2008reciprocal, van2011reciprocal, zhou2017fast, wang2017safety}. Discrete graph-based methods can generate optimal paths to hundreds of robot but has no guarantee of dynamic feasibility \cite{yu2015pebble, yu2016optimal, wagner2015subdimensional, sharon2015conflict, wagner2011m}. The solution paths can be transformed to dynamically feasible trajectories through post-processing methods, but is not adequate to handle robots that are non-holonomic or have drifts \cite{honig2016multi, honig2018trajectory, park2020efficient, tang2018hold, bandyopadhyay2017distributed, turpin2014capt}. Decoupled approaches are computationally efficient and generate dynamically feasible trajectories, but have no optimality guarantees as they avoid planning in the joint-space \cite{vcap2015prioritized, robinson2018efficient, park2020efficient, morgan2014model}. Sampling-based methods, in theory, have properties of asymptotical optimality and completeness but the rate of convergence is extremely slow and far from practical \cite{vsvestka1998coordinated, sanchez2002using, vcap2013multi, dobson2017scalable, solovey2016finding}.



The proposed approach in this paper tackles MRMP fully in the continuous joint-space with dynamics considered. Instead of relying on simplifications, it leverages upon a recently developed parabolic relaxation that is fundamentally more efficient than the state-of-the-art convexification methods due to its reliance on convex quadratic constraints as opposed to conic constraints. In effect, our method can efficiently generate dynamically feasible and optimal trajectories to MRMP. The overview of the approach is shown in Fig. 1. First, multi-robot motion planning problem is mathematically formulated as a non-convex optimization. In this original formulation, non-convexity exists in the collision avoidance constraints. Then the formulation is lifted into a higher dimensional space where all the non-convexity is captured in a single constraint involving an auxiliary matrix introduced. From there, parabolic relaxation is performed to efficiently convexify the formulation and a penalty term is incorporated into the objective function to ensure feasibility. The aforementioned lifted, convexified, and penalized formulation is solved sequentially with solution update at each round till convergence.

Our main contributions and novelty are: i) to our knowledge, this is first approach that is computationally tractable in finding optimal and dynamical feasible trajectories in the joint-space for MRMP, ii) we introduce a computationally-efficient convex relaxation for MRMP that does not rely on local approximation that most of the state-of-the-art approaches take, iii) a penalization technique is developed to ensure dynamical feasibility, and iv) the proposed method is applied sequentially in order to obtain feasible and high-quality solutions for MRMP.





The paper is organized as follow: In Section \ref{rel_work}, we give an overview on related previous work. In Section \ref{prob_formulation}, we present mathematical formulation of the problem with convex quadratic constraints with description of the notations used. In Section \ref{alg_des}, we elaborate our multi-stage optimization approach with details of each step that involve lifting, convexification, and penalization with iterative search algorithm. In Section \ref{experiments}, we show experimental results for the approach on different scenarios. We provide a discussion of the advantages and shortcomings of our approach in Section \ref{conclusion} and conclude with possible future research directions.

\subsection{Related Work}\label{rel_work}
Multi-robot motion planning (MRMP) problems have been actively studied in varied formulations over the past decades. MRMP is inherently challenging due to high dimensionality and non-convexity involved, and many different approaches exist in the literature to tackle it. We present state-of-the-art methods and discuss pros and cons of each method in relation to our approach.


Local methods reactively plan motion for each robot by considering positions and velocities of other robots in its neighborhood \cite{guy2009clearpath, lalish2012distributed, van2008reciprocal, van2011reciprocal, zhou2017fast, wang2017safety}. The main focus is on how to guarantee collision avoidance among the robots and Optimal Reciprocal Collision Avoidance (ORCA) \cite{van2008reciprocal} is a well-known algorithm in this approach. These methods are computationally efficient, can scale to a very large number of robots, and practical to real-time applications. However, the local and reactive nature that makes it efficient, also leads to its limitations. These methods are highly susceptible to local minima and the resulting robot motions are far from optimal. It may also lead to a deadlock and hence does not guarantee that robots will reach given goals. 


Path planning methods focus on generating path for each robot with no consideration to robot's dynamics \cite{yu2015pebble, yu2016optimal, wagner2015subdimensional, sharon2015conflict, wagner2011m}. These methods often confine robots to a graph where a robot is allowed to move from one vertex to its adjacent vertex at each time step. Then graph search techniques are utilized to find collision-free path along the graph to transition each robot from its current to goal vertex. As robots are confined to a given graph, this approach has benefit of finding feasible solution in polynomial time of $O(n^3)$ \cite{yu2015pebble}. Computing optimal paths also becomes computationally tractable, and with effective heuristics, near-optimal path for hundreds of robots can be found within seconds \cite{yu2016optimal}. However, the solution paths do not guarantee dynamical feasibility and robots may not be able to follow it, especially if they are non-holonomic or have drifts. Moreover, the optimality of paths on a discrete graph does not translate to optimal path in the continuous state-space of robot and may actually be far from optimal.



Decoupled methods avoid high dimensionality of the joint-space by planning trajectories of each robot independently. \cite{vcap2015prioritized, robinson2018efficient, park2020efficient, morgan2014model}. Priorities are given to each robot and trajectories are planned sequentially from the highest to the lowest, with a robot avoiding higher priority robots at each iteration. As planning is performed in the state-space of a single robot, optimization techniques can be used to efficiently generate trajectories that satisfy dynamic feasibility and other user-defined constraints. Also, these methods enjoy linear scaling with respect to the number of robots. However, this approach suffer from incompleteness and tend to fail in a crowded environment. Moreover, the optimality of each trajectory in sequence does not lead to optimality in the joint-space.

Recent works combine path planning with optimization-based trajectory generation to take advantage of both approaches \cite{honig2016multi, honig2018trajectory, park2020efficient, tang2018hold, bandyopadhyay2017distributed, turpin2014capt}. These methods operate in two phases. In the first phase, path planning methods are utilized to efficiently generate collision-free geometric paths. Then the resulting paths are post-processed through optimization techniques, to generate refined smooth trajectories that guarantee dynamical feasibility. This combination alleviates the problem of dynamic infeasibility of path planning and computational intractability of trajectory planning in joint-space that each approach has. It has been demonstrated to generate collision-free smooth trajectories for hundreds of quadrotors in dense environment within a few minutes \cite{honig2018trajectory}. However, abstracting away dynamics to generate paths has its limitations. These methods are not easily extendable to robots that are non-holonomic or have drifts, and optimality is not guaranteed.

Our approach solve the MRMP by planning trajectories through optimization directly in the continuous joint-space of the all robots, which has been deemed computationally intractable. Previous works \cite{augugliaro2012generation, mellinger2012mixed} take similar approach but has limited scalability and only been demonstrated on a few robots. Several works extend sampling-based methods from single to multi-robot, but suffer from extremely slow convergence despite theoretical guarantee of asymptotical optimality \cite{vsvestka1998coordinated, sanchez2002using, vcap2013multi, dobson2017scalable, solovey2016finding}. Our method tackles the computational complexity of planning in the joint-space by leveraging recently developed Parabolic relaxation \cite{madani2017low, kheirandishfard2018convex, kheirandishfard2018convex2}. Since it does not rely on discretizing or decoupling the state-space, trajectories generated are both optimal and dynamically feasible. Further, it is highly scalable and can generate optimal trajectories for a hundred robots in an extremely cluttered environment within several minutes.

\section{Problem Formulation}\label{prob_formulation}

Given a set of robots and obstacles $\Ncal\triangleq\{1,2,\ldots, |\Ncal|\}$ and a planning horizon $\Tcal\triangleq\{1,2,\ldots,|\Tcal|\}$, we formalize the problem of multi-robot motion planning (MRMP) as a trajectory planning problem in the joint state-space of all the members of $\Ncal$. Define $\Rcal\subseteq\Ncal$ as the set of robots. We assume that each robot is operating in an $n$-dimensional configuration space $\Rbb^n$ with the corresponding state-space $\Rbb^{2n}$. For each $i\in\Ncal$ and $t\in\Tcal\cup\{|\Tcal|+1\}$, the state of the robot or obstacle $i$ at time $t$ is represented by $\xbf_i[t]\in\Rbb^{2n}$. For each $i\in\Ncal$ define $r_i$ as the collision radius of the robot or obstacle $i$. We define the matrix $\Gbf\triangleq[\Ibf_{n\times n} \;\; \boldsymbol{0}_{n\times n}]$ that can be multiplied to retrieve position from a state vector. The control input of each robot at time $t$ is denoted by $\{\ubf_i[t]\in\Rbb^m\}_{i\in \Ncal}$, where $m = n$ if the robot is fully actuated and $m=0$ for obstacles. The joint state-space of all $|\Ncal|$ robots and obstacles at time $t$ is denoted by $\Xbf[t]\triangleq\left[\xbf_1[t],\ldots,\xbf_{|\Ncal|}[t]\right]$.

Given the above definitions, the proposed formulation of the MRMP problem is as follows:
\begin{subequations} \label{qcqp_original}
\begin{align}
&\!\!\mathrm{minimize}\quad\sum_{t\in \Tcal}^{~}\sum_{i\in \Rcal}^{~}{{\|\ubf_i[t]\|_q}
}\label{objective}\\
&\!\!\mathrm{subject~to}\nonumber\\
&\xbf_i[t+1]=\Abf_i \xbf_i[t]+\Bbf_i \ubf_i[t]
&&\hspace{-0mm}\forall i\!\in\! \Rcal,\; \forall t\! \in\! \Tcal \label{dynamics}\\
&\xbf_i[1]=\xbf_i^{\mathrm{init}},\;\xbf_i[|\Tcal|\!+\!1]=\xbf_i^{\mathrm{goal}}
&&\hspace{-0mm}\forall i\!\in\! \Rcal \label{init_goal_state}\\
&\|\ubf_i[t]\|_p\leq u^{\max}
&&\hspace{-0mm}\forall i\!\in\! \Rcal,\;\forall t\!\in\! \Tcal \label{control_bound}\\
&\|\Gbf(\xbf_i[t]-\xbf_j[t])\|_2\geq r_i+r_j
&&\hspace{-0mm}\forall i\!\neq\! j\!\in\! \Ncal,\;\forall t\!\in\! \Tcal\!\! \label{robot_avoid}
\end{align}
\end{subequations}
where $p,q\in[1,2]$ are constant parameters. The objective function in Eq. (\ref{objective}) has control effort as the first term to minimize the actuation cost. 
Dynamical feasibility is enforced by Eq. (\ref{dynamics}), where we use a linearized model of robots' dynamics for simplicity. For heterogeneous system, the matrices 
${\Abf_i\in\Rbb^{2n\times 2n}}$ and 
${\Bbf_i\in\Rbb^{2n\times m}}$ will vary across the robots, but remain constant for homogeneous system. Note that robots' dynamics need not be confined to linear systems. One can use an alternative quadratic form to represent dynamics of non-holonomic robots (e.g. wheeled mobile robot) with careful manipulation. Collision avoidance of each robot with obstacles and other robots is imposed by Eq. (\ref{robot_avoid}) which are non-convex constraints. In this formulation, we use simple bounding spheres for both robots and obstacles with radius $r_i$ to check for collisions. More elaborative bounding shape can be used if necessary.
The formulation is labeled trajectory planning where start and goal states are set as in Eq. (\ref{init_goal_state}), and cannot be interchanged among robots. The bounds on control input is given in Eq. (\ref{control_bound}). Note that we assume the trajectories of obstacles are known prior and not part of the decision variables. 

The non-convex formulation of Eq. (\ref{qcqp_original}) will be referred to as the original problem hereinafter throughout the paper.

\section{Algorithm Description}\label{alg_des}

In the original problem of Eq. (\ref{qcqp_original}), non-convexity exists in the collision-avoidance constraints. To effectively deal with the non-convexity, we first lift the original problem into a higher dimensional space. In the lifted reformulation, all the non-convexity of the original problem is captured solely by the auxiliary matrix being introduced. From there, we perform parabolic relaxation to convexify the auxiliary matrix in a computationally efficient manner. Furthermore, a penalty term is incorporated into the objective function to achieve feasibility of for the original problem. The aforementioned lifted, convexified, and penalized formulation is solved sequentially with solution update at each iteration till convergence stopping criteria are met.



\subsection{Lifted Formulation}

We lift the original problem of Eq. (\ref{qcqp_original}) into a higher dimensional space by introducing an auxiliary matrix $\Ybf[t]\in\Sbb_n$ for every $t \in \Tcal$. $\Ybf[t]$ is a $n\times n$ symmetric matrix  whose $(i,j)$ element accounts for the inner product $\xbf_i[t]^\top\Gbf^{\top}\Gbf\xbf_j[t]$. This serves to capture all the non-convexity of the original problem solely by the auxiliary matrices introduced. In effect, the non-convex constraints of the original problem become \emph{linear} in the lifted space. Here is the lifted formulation with respect to the new variable $\Ybf[t]$:



\begin{subequations}
\begin{align}
&\!\!\mathrm{minimize}\quad \sum_{t\in \Tcal}^{~}\sum_{i\in \Rcal}^{~}{{\|\ubf_i[t]\|_q}
}\label{obj_lifted}\\
&\!\!\mathrm{subject~to}\nonumber\\
& \xbf_i[t+1]=\Abf_i \xbf_i[t]+\Bbf_i \ubf_i[t]
&&\hspace{-0mm}\forall i\!\in\! \Rcal,\; \forall t\! \in\! \Tcal\\
& \xbf_i[1]=\xbf_i^{\mathrm{init}},\;\xbf_i[|\Tcal|+1]=\xbf_i^{\mathrm{goal}}
&&\hspace{-0mm}\forall i\!\in\! \Rcal\\
& \|\ubf_i[t]\|_p\leq u^{\max}
&&\hspace{-0mm}\forall i\!\in\! \Rcal,\;\forall t\!\in\! \Tcal\\
& Y_{ii}[t]\!+\!Y_{jj}[t]\!-\!2Y_{ij}[t]\geq (r_i + r_j)^2  
&&\hspace{-0mm}\forall i\!\neq\! j\!\in\! \Ncal,\;\forall t\!\in\! \Tcal\!\! \label{robot_avoid_lifted_1} \\
& Y_{ij}[t]=\xbf_i[t]\T\Gbf^{\top}\Gbf\xbf_j[t] 
&&\hspace{-0mm}\forall i,j\!\in\! \Ncal,\;\forall t\!\in\! \Tcal\!\!\label{aux_matrix} 
\end{align}
\end{subequations}

The collision avoidance constraints Eq. (\ref{robot_avoid_lifted_1}) are now linear in the lifted formulation. Note that this is a \emph{lossless} transformation - solving the lifted formulation exactly recovers the optimal solution to the original problem. The non-convexity of the original problem is captured by the auxiliary matrix $\Ybf[t]$ of Eq. (\ref{aux_matrix}) that we tackle in the next step.

\subsection{Convexification}
We proceed to handle the non-convexity of $\Ybf[t]$ through convexification. The state-of-the-art and common practice is semi-definite programming (SDP) relaxation. Using SDP relaxation, the Eq. (\ref{aux_matrix}) can be convexified to the following.

\vspace{-3mm}
{\small
\begin{align}
\Ybf[t]\!\succeq\!\left[\xbf_1[t],\ldots,\xbf_{|\Ncal|}[t]\right]^{\!\top}
\!\Gbf^{\!\top}\Gbf
\left[\xbf_1[t],\ldots,\xbf_{|\Ncal|}[t]\right] \;\; \forall t\!\in\! \Tcal\!	\label{sdp}
\end{align}}
According to Schur complement Eq. \eqref{sdp} is equivalent to:
{\small
\begin{align}
\!\!\Bigg[\begin{matrix}
\Ybf[t] & \big[\xbf_1[t],\ldots,\xbf_{|\Ncal|}[t]\big]^{\!\top}\Gbf^{\!\top}\\
\Gbf\big[\xbf_1[t],\ldots,\xbf_{|\Ncal|}[t]\big] & \Ibf_{n\times n}
\end{matrix}\Bigg]\!\succeq\! 0\;\;\forall t\!\in\! \Tcal\!\! \label{sdp_schur}
\end{align}}

However, SDP relaxation can be computationally demanding and 
may not scale well to large-scale matrices \cite{zhang2017sparse}. Instead, we leverage on the recently developed parabolic relaxation to yield a low-complexity convex formulation.

\subsubsection{Parabolic Relaxation}
Parabolic relaxation provides computationally efficient low-complexity convexification that is orders-of-magnitude faster than the state-of-the-art SDP relaxation. Its mechanism and efficiency is detailed in \cite{madani2017low}. Following parabolic relaxation, Eq. \eqref{aux_matrix} is convexified as:

\vspace{-3mm}
{\small
\begin{subequations}
\begin{align}
&\!\!Y_{ii}[t]\!+\!Y_{jj}[t]\!+\!2Y_{ij}[t]\!\geq\!\|\Gbf(\xbf_j[t]\!+\!\xbf_i[t])\|^2_2\;\;\forall i, j\!\in\! \Ncal,\forall t\!\in\! \Tcal\!\!\label{para1}\\
&\!\!Y_{ii}[t]\!+\!Y_{jj}[t]\!-\!2Y_{ij}[t]\!\geq\!\|\Gbf(\xbf_j[t]\!-\!\xbf_i[t])\|^2_2\;\;\forall i, j\!\in\! \Ncal,\forall t\!\in\! \Tcal\!\!\label{para2}
\end{align} 
\end{subequations}}
%
It can be easily observed that there exists an $Y_{ij}[t]$ that satisfies Eq. \eqref{para1}, \eqref{para2} and \eqref{robot_avoid_lifted_1}, if and only if:
\begin{subequations}\label{para_simple}
\begin{align}
&\!\!\! Y_{ii}[t]\geq\|\Gbf(\xbf_i[t]\|^2_2 
&&\hspace{-29mm} \forall i\!\in\! \Ncal,\;\forall t\!\in\! \Tcal\\
&\!\!\! 2(Y_{ii}[t]+Y_{jj}[t])\geq(r_i+r_j)^2+\|\Gbf(\xbf_j[t]+\xbf_i[t])\|^2_2 \nonumber \\
&\!\!\!
&&\hspace{-29mm} \forall i\!\neq\! j\!\in\! \Ncal,\;\forall t\!\in\! \Tcal
\end{align}
\end{subequations}
This allows us to omit the off-diagonal elements $Y_{ij}[t]$, and substitute the constraints \eqref{para1}, \eqref{para2} and \eqref{robot_avoid_lifted_1} with the above, which helps with the scalability as these are $\frac{|\Tcal||\Ncal|(|\Ncal|-1)}{2}$ additional variables that are being reduced. We utilize Eq. \eqref{para_simple} as convexification of Eq. \eqref{aux_matrix} in our final form.

\begin{subequations}
\begin{align}
&\!\!\!\!\!\mathrm{minimize}\;\; \sum_{t\in \Tcal}^{~}\sum_{i\in \Rcal}^{~}{{\|\ubf_i[t]\|_q}} \\
&\!\!\!\!\!\mathrm{subject~to}\nonumber\\
&\!\!\!\xbf_i[t+1]=\Abf_i \xbf_i[t]+\Bbf_i \ubf_i[t]
&&\hspace{-29mm} \forall i\!\in\! \Rcal,\; \forall t\! \in\! \Tcal\\
&\!\!\!\xbf_i[1]=\xbf_i^{\mathrm{init}},\;\xbf_i[|\Tcal|+1]=\xbf_i^{\mathrm{goal}}
&&\hspace{-29mm} \forall i\!\in\! \Rcal\\
&\!\!\! \|\ubf_i[t]\|_p\leq u^{\max}
&&\hspace{-29mm} \forall i\!\in\! \Rcal,\;\forall t\!\in\! \Tcal\\
&\!\!\! Y_{ii}[t]\geq\|\Gbf(\xbf_i[t]\|^2_2 
&&\hspace{-29mm} \forall i\!\in\! \Ncal,\;\forall t\!\in\! \Tcal\\
&\!\!\! 2(Y_{ii}[t]+Y_{jj}[t])\geq(r_i+r_j)^2+\|\Gbf(\xbf_j[t]+\xbf_i[t])\|^2_2 \nonumber \\
&\!\!\!
&&\hspace{-29mm} \forall i\!\neq\! j\!\in\! \Ncal,\;\forall t\!\in\! \Tcal
\end{align}
\end{subequations}
Notice that the solution of the aforementioned relaxations are not necessarily feasible for Eq. \eqref{qcqp_original}. In the following section, we propose to revise the objective function to direct convex relaxations towards finding feasible points for the original non-convex problem \eqref{qcqp_original}.




\subsection{Penalization}
In this section, we incorporate a penalty term into the objective function to achieve feasibility of the solution from the convexified formulation to the original problem. The penalty term introduced is:
\begin{align}
\eta\times\sum_{t\in\Tcal}^{~}\sum_{i\in\Ncal}{
Y_{ii}[t]-2\check{\xbf}_i[t]^{\!\top}\Gbf^{\!\top}\Gbf\xbf_i[t]}
\end{align}
where $\check{\xbf}_i[t]$ is an initial guess for the solution and $\eta$  is a fixed parameter. The lifted, convexified, and penalized formulation of the original problem Eq . \eqref{qcqp_original} is as follow:
\begin{subequations}\label{qcqp_convexified}
\begin{align}
&\!\!\mathrm{minimize}\nonumber\\ 
&\sum_{t\in \Tcal}^{~}\!\sum_{i\in \Rcal}^{~}\!{\|\ubf_i[t]\|_q}
+
\eta\sum_{t\in\Tcal}^{~}\sum_{i\in\Ncal}{
Y_{ii}[t]-2\check{\xbf}_i[t]^{\!\top}\!\Gbf^{\!\top}\!\Gbf\xbf_i[t]}
\\
&\!\!\mathrm{subject~to}\nonumber\\
&\xbf_i[t+1]=\Abf_i \xbf_i[t]+\Bbf_i \ubf_i[t]
&&\hspace{-30mm} \forall i\!\in\! \Rcal,\; \forall t\!\in\! \Tcal\\
&\xbf_i[0]=\xbf_i^{\mathrm{init}},\;\xbf_i[|\Tcal|]=\xbf_i^{\mathrm{goal}}
&&\hspace{-30mm} \forall i\!\in\! \Rcal\\
& \|\ubf_i[t]\|_p\leq u^{\max}
&&\hspace{-30mm} \forall i\!\in\! \Rcal,\;\forall t\!\in\! \Tcal\\
& Y_{ii}[t]\geq\|\Gbf\xbf_i[t]\|^2_2 
&&\hspace{-30mm} \forall i\!\in\! \Ncal,\;\forall t\!\in\! \Tcal\\
& 2(Y_{ii}[t]+Y_{jj}[t])\geq(r_i+r_j)^2+\|\Gbf(\xbf_j[t]+\xbf_i[t])\|^2_2 \nonumber \\
&&&\hspace{-30mm} \forall i\!\neq\! j\!\in\! \Ncal,\;\forall t\!\in\! \Tcal
\end{align}
\end{subequations}

The next theorem investigates conditions under which the penalized convex relaxation problem \eqref{qcqp_convexified} with a feasible initial point leads to a feasible point for the original non-convex formulation \eqref{qcqp_original}.

\begin{theorem}
	[Feasibility preserving]\label{thm1} Let the pair 
	$\{\check{\boldsymbol{X}}[t]\}_{t\in \Tcal}$ and 
	$\{\check{\boldsymbol{u}}_i[t]\}_{i\in\Rcal,t\in \Tcal}$ 
	represent a feasible initial point that satisfies the linear independence constraint qualification (LICQ) condition for constraints \eqref{dynamics} - \eqref{robot_avoid}. If $\eta$ is sufficiently large, then the penalized convex relaxation Eq. \eqref{qcqp_convexified} has a unique optimal solution
	$\{\accentset{*}{\boldsymbol{X}}[t]\}_{t\in \Tcal}$ and 
	$\{\accentset{*}{\boldsymbol{u}}_i[t]\}_{i\in\Rcal,t\in \Tcal}$ 
	which is feasible for the original nonconvex problem \eqref{qcqp_original}, and satisfies:
	\begin{align}
	\sum_{t\in \Tcal}^{~}\!\sum_{i\in \Rcal}^{~}\!{\|\accentset{*}{\ubf}_i[t]\|_q}\leq
	\sum_{t\in \Tcal}^{~}\!\sum_{i\in \Rcal}^{~}\!{\|\check{\ubf}_i[t]\|_q}
	\end{align}
\end{theorem}

\begin{proof}
Please see \cite{kheirandishfard2018convex} for the proof.
\end{proof}

According to Theorem \ref{thm1}, the penalized convex relaxation
\eqref{qcqp_convexified} preserves the feasibility of an initial point. In \cite{kheirandishfard2018convex}
it is shown that even if the initial point is not
feasible, but sufficiently close to the feasible set,
the penalized convex relaxation is still guaranteed to
provide a feasible point.

\subsection{Sequential Penalization} 
Motivated by Theorem \ref{thm1}, this section presents
a sequential approach that solves a sequence of penalized
relaxations of the form \eqref{qcqp_convexified} to infer high-quality feasible
points for the non-convex problem \eqref{qcqp_original}. The proposed
scheme starts from an arbitrary initial point. Once a
feasible point for \eqref{qcqp_original} is obtained, according to Theorem
\ref{thm1}, the proposed scheme preserves feasibility and generates
a sequence of points whose objective values monotonically
improve. The details of this sequential approach are delineated
in Algorithm 1. The following theorem guarantees the convergence
of Algorithm 1 to at least a locally optimal solution.

\begin{algorithm}[t]
	\caption{Sequential Penalized Relaxation}
	\label{SP-SDP}
	\begin{algorithmic}[1]
		\Require {$\{\check{\boldsymbol{X}}[t]\}_{t\in \Tcal}$, a fixed parameter $\eta>0$.}
		\State $k\leftarrow 0$ 
		\Repeat
		\State $k\leftarrow k+1$ 
		\State $\{\boldsymbol{X}^k[t]\}_{t\in \Tcal}\leftarrow$ solve the penalized relaxation \eqref{qcqp_convexified}.
		\State $\{\check{\boldsymbol{X}}[t]\}_{t\in \Tcal}\leftarrow \{\boldsymbol{X}^k[t]\}_{t\in \Tcal}$ 
		\Until stopping criteria is met
		\Ensure {$\{\boldsymbol{X}^k[t]\}_{t\in \Tcal}$}
	\end{algorithmic}
\end{algorithm}

\begin{theorem}	[Convergence]\label{thm2}
Let $\check{\mathcal{F}}$ denote a bounded epigraph of the problem \eqref{qcqp_original}, whose every member satisfies the constraint qualification (LICQ) condition for constraints \eqref{dynamics} - \eqref{robot_avoid}. If $\eta$ is sufficiently large, then the sequence generated by Algorithm 1 converges to a point that satisfies the Karush–Kuhn–Tucker optimality conditions for \eqref{qcqp_original}.
\end{theorem}

\begin{proof}
	Please see \cite{kheirandishfard2018convex} for the proof.
\end{proof}

\begin{figure*}[t!]
	\centering
	\includegraphics[width=1.0\linewidth]{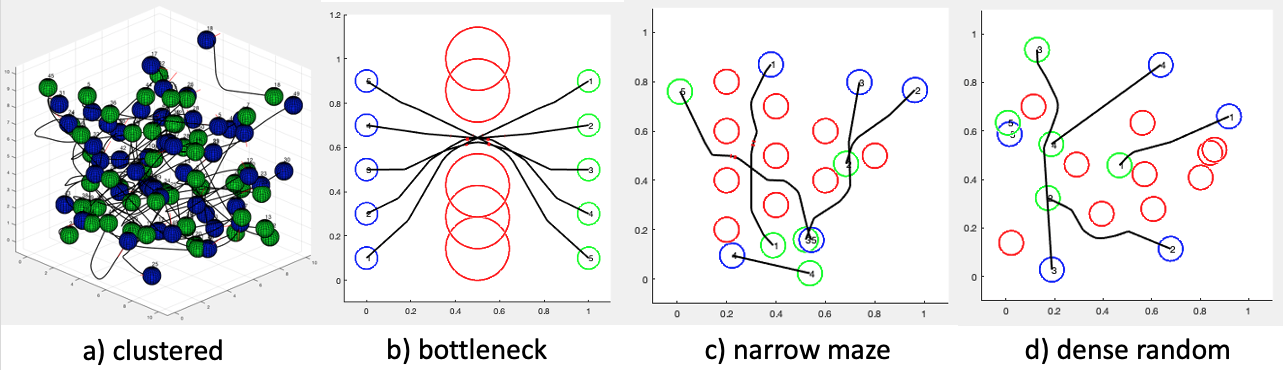}
	\caption{Our approach is able to handle and generate optimal solutions for difficult problem instances. The plots show fuel optimal trajectories generated by our approach for a) extremely clustered 3D environment with large number of agents, b) agents converging towards a bottleneck, c) navigating through narrow maze, and d) densely packed map with random obstacles and agents positions. Robots' initial and goal positions are marked in blue and green respectively. Obstacles are shown in red. The state-of-the-art methods either return sub-optimal solutions or become infeasible for these problem instances.}
	\label{fig:challenge_prob}
\end{figure*}

\begin{figure}[t]
	\centering
	\includegraphics[width=1.0\columnwidth]{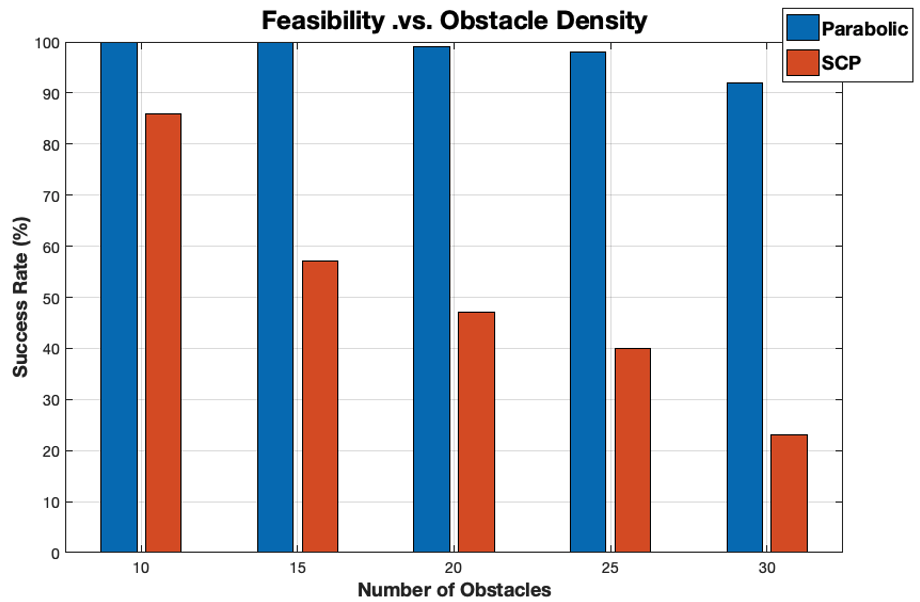}
	\caption{MRMP via parabolic has significantly higher success rate compared to the state-of-the-art methods as it plans in the join-space without local approximation. Randomized map with varying obstacles density is used as the unit test for quantifying the success rate.}
	\label{fig:feas_vs_obs}
\end{figure}

\section{Experiments}\label{experiments}
We present results from simulation experiments comparing our proposed approach with state-of-the-art baselines. We show that MRMP via Parabolic can effectively handle challenging environments that are extremely dense for which the state-of-the-art methods either become infeasible or return solutions of poor quality. We demonstrate that our approach is computationally tractable for swarm, despite planning in the joint-space. We further illustrate how our approach can recover from a bad initial seed and iteratively converges towards the optimal point. The supplemental video has animation of optimal trajectories generated with additional simulations performed.

\subsection{Performance on Challenging Environment}
MRMP via parabolic plans in the full joint-space of all robots and does not rely on local approximation. This enables our approach to solve difficult problem instances that are challenging for motion planners to handle. These include a) extremely clustered environment with large number of agents, b) agents simultaneously converging towards a bottleneck to reach the goal, c) navigating through narrow maze with low clearance, and d) densely packed map with randomized obstacles and agents positions.

Figure \ref{fig:challenge_prob} shows fuel optimal trajectories generated by our approach for the aforementioned problem instances. Here, robots are following 3D or 2D double integrator dynamics with the assumption of thrusters in each of its axis. The objective cost to minimize is summation of $\ell_1$ norm ($q=1$ and $p=1$) of control effort over all robots. We used Algorithm 1 with initial seeds that are linear interpolation of lines connecting robots' initial and goal states. The simulations are run with $\eta=50$.


The state-of-art methods either return sub-optimal solutions or become infeasible for the examples shown in Fig. \ref{fig:challenge_prob}. ORCA \cite{van2008reciprocal} is able to return feasible solutions for these problems, but has very high fuel cost as it locally reacts to avoid collisions and has no notion of cost. Sampling-based mRRT* \cite{vcap2013multi} has extremely slow convergence rate that is not practical for these problems. SCP methods \cite{morgan2014model} return optimal solutions similar to our Parabolic approach, but often become infeasible when environments become highly dense.


To quantify the success rate of both MRMP via parabolic and SCP that return feasible solutions, we run simulation experiments on randomized problems instances of varying obstacle density. In an arena of dimension 1m x 1m, we randomly place both robots and obstacles of identical circular shape with diameter of 0.1m each. The number of robots is fixed to five, and number of obstacles is varied from 10 to 30. The goal for each robot is also randomly chosen. Figure \ref{fig:challenge_prob}d) is showing a problem instance with 10 obstacles. The robot is following 2D double integrator dynamics and objective is to minimize the summation $\ell_1$ norm of control effort over all robots  ($q=1$ and $p=1$).  For all trials we have set $\eta=50$ and used termination criteria that stops when relative delta between objective costs of two successive steps are within 0.01\% for Algorithm 1 and SCP.
Figure \ref{fig:feas_vs_obs} shows success rate of our approach in comparison to SCP measured over 100 randomized problem instances for each number of obstacles. As success rate of optimization-based methods is highly affected by an initial seed, a linear interpolation from start to goal is given to both Parabolic and SCP for a fair comparison. The result shows SCP has diminishing success rate as obstacle density increase, returning feasible solution less than 25\% of randomized problem instances with 30 obstacles. Our approach, on the other hand, returns feasible solution even for highly dense situations, with more than 90\% success rate even for randomized problem instances with 30 obstacles. This is mainly because MRMP via parabolic plans in the full joint-space of all robots and does not suffer issues that comes with relying on local approximations. However, as a trade-off, MRMP via Parabolic has higher run-time to convergence compared to SCP that gains computational efficiency through local approximation. 

\begin{figure*}[t!]
	\centering
	\includegraphics[width=1.0\linewidth]{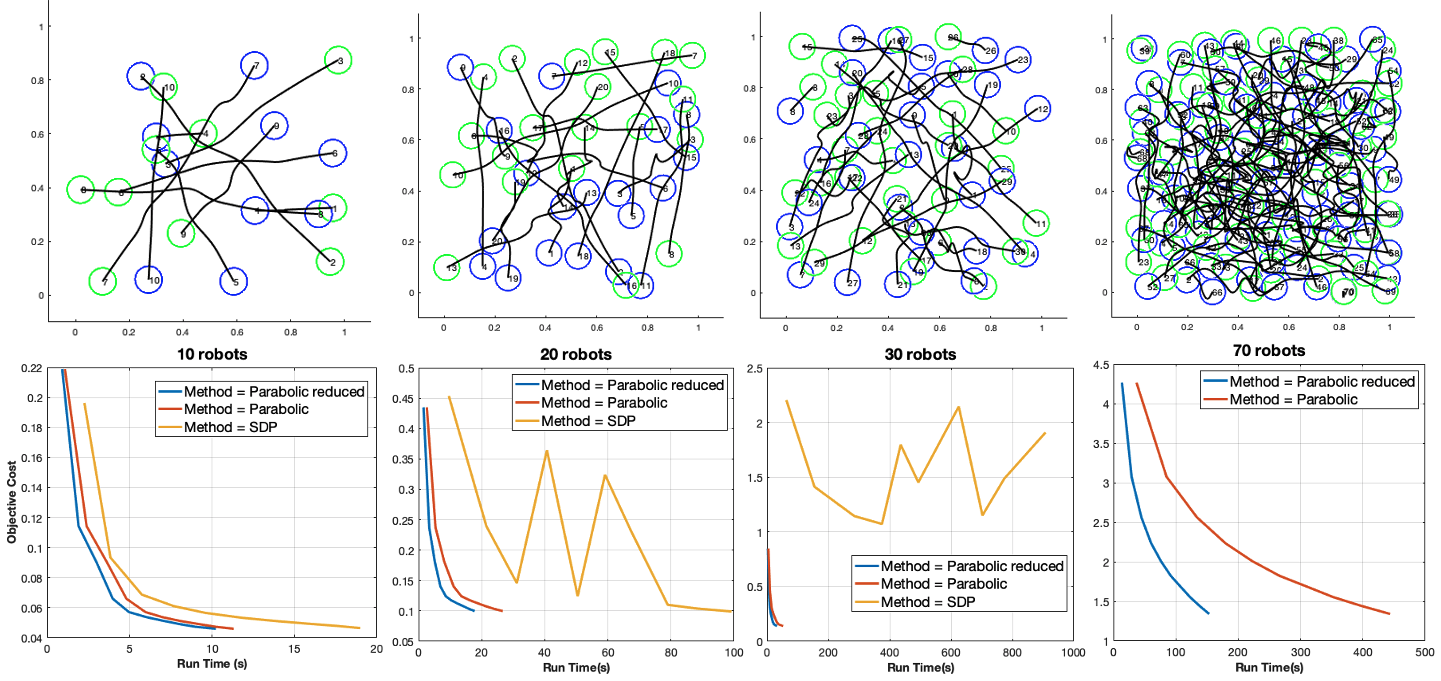}
	\caption{Our approach makes planning in full joint-space of all robots computationally tractable for large number of robots, by leverage upon advances in Parabolic relaxation [34]-[36]. The state-of-the-art and commonly used Semi-definite programming (SDP) relaxation is computationally expensive and fails to converge as the size of robot fleet grows. MRMP via Parabolic is computationally efficient and tractable even for swarm of robots.}
	\label{fig:tractability}
\end{figure*}

\begin{figure}[t!]
	\centering
	\includegraphics[width=1.0\columnwidth]{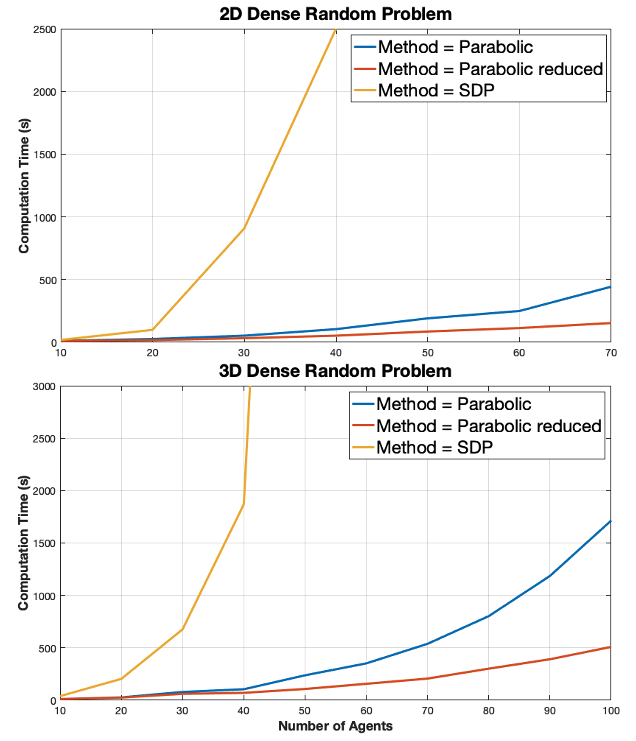}
	\caption{MRMP via parabolic remains computationally tractable for up to 100 robots for both 2D and 3D problems despite inherent exponential run-time associated with MRMP in joint-space that is NP-hard. The state-of-the-art SDP relaxation quickly becomes computationally intractable as the number of robots become larger than 30.
	}
	\label{fig:run_time}
\end{figure}

\subsection{Computational Tractability}
Planning in full joint-space is computationally challenging due to high dimensionalilty of the state-space involving all robots. MRMP in joint-space cannot avoid the curse of dimensionality as it is shown to be PSPACE-hard even for simple robot geometry \cite{solovey2016hardness}. However, the exponential growth in run-time can be kept tractable up to a large number of robots by leveraging upon the proposed approach.

To demonstrate computational tractability of MRMP via parabolic, we run simulation experiments on both 2D and 3D randomized problems of varying number of robots. Similar to the previous experiment that quantified the success rate, we randomly place robots of identical circular shape with diameter of 0.1m inside an arena of dimension 1m x 1m. We start with 10 robots and increase the number of robots up to 100 at an incremental of 10. The 3D problems are identical but with spherical shape robots and cubical arena of unit dimension. Robots follow 2D or 3D double integrator dynamics and objective is to minimize the summation L1 norm of control effort over all robots. Each problem is solved using SDP relaxation in Eq. \eqref{sdp_schur}, parabolic relaxation in Eq. (5a), (5b), and simplified Parabolic relaxation in Eq. \eqref{para_simple}, respectively for comparison.

Figure \ref{fig:tractability} shows fuel optimal trajectories generated for different number of robots in 2D (above) with plot of how objective cost converges with respect to run-time for each case (below). The computational benefit of using parabolic can be seen even from low-cardinality swarm of 10 robots in comparison with the state-of-the-art SDP. SDP starts to have numerical issues when the number of robots is increased to 20 and fails to converge for 30 robots. MRMP via Parabolic, on the other hand, is computationally tractable and converges for the randomized 2D problem with high-cardinality swarm of 70 robots. Note that the randomized 2D problem cannot be generated without robots overlapping each other for 80 robots and above.

Figure \ref{fig:run_time} shows run-time of each method as the number of robots increase for both 2D and 3D randomized problems described above. Using SDP relaxation, the computation time quickly become intractable with number of robots growing beyond 30. In comparison, MRMP via Parabolic remains computationally tractable even for swarm of 100 robots despite exponential growth that is inherently unavoidable for MRMP in joint-space. The run-time reported is based on a prototype implementation of the algorithms in MATLAB using CVX and running SDPT3 as the back-end solver on a laptop with 3.1 GHz Intel Core i7. This can be further reduced if it is implemented in C/C++ and the code is optimized for performance. Our numerical algorithm is also compatible with both GPU and GPU and can be parallelized to have orders-of-magnitude reduction in the run-time from the reported values.

\begin{figure}[t]
	\centering
	\includegraphics[width=1.0\columnwidth]{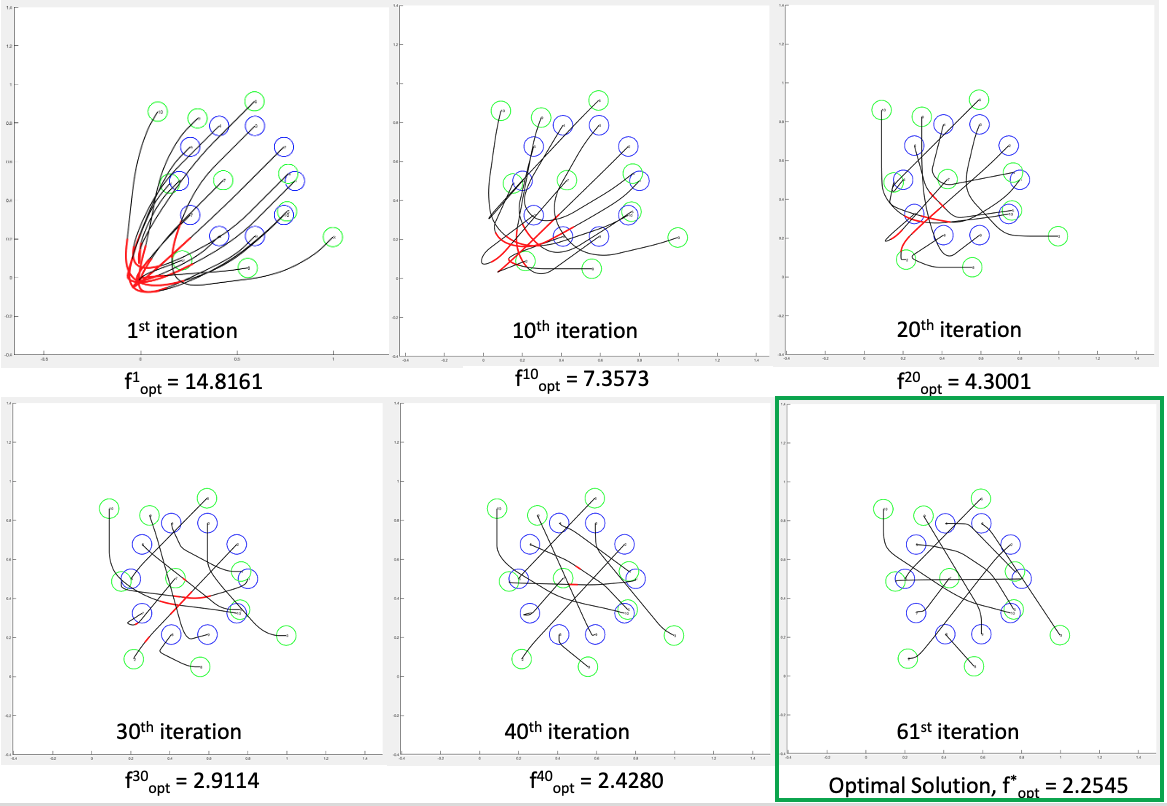}
	\caption{Our approach iteratively solves through lifted, convexified, and penalized formulation in Eq. \eqref{qcqp_convexified} to converge to optimal solution of the original problem in Eq. \eqref{qcqp_original}. Through iterations, MRMP via Parabolic can reach optimal solution even when a bad initial seed is given as shown. The computed trajectories are plotted at iterations and convergence to optimum is reached at 61st iteration for the shown problem instance.}
	\label{fig:iteration}
\end{figure}

\subsection{Iterative Convergence to Optimum}
Our approach reaches optimality solution to the original problem in Eq. \eqref{qcqp_original} by sequentially solving the lifted, convexified, and penalized formulation in Eq. \eqref{qcqp_convexified} with updates to the penalty term at each iteration. Starting from a given initial seed, MRMP via parabolic will gradually improve upon its computed trajectories at each iteration as shown in Fig. \ref{fig:iteration}. The rate of convergence will be dependent on the quality of the initial seed given. If the initial seed is close to the optimum, it will only need few iteration for convergence, but a poor seed will require a large number of iterations and may even fail to converge. Nevertheless, through iterative revision of the penalty term, our approach is able to converge to the optimal solution even when it is provided with a poor initial seed. The example shown in Fig. \ref{fig:iteration} illustrates how MRPP via parabolic is able to recover from a bad initial seed that forced all robots to collide at left bottom corner of the arena. The plot shows segments of trajectories in collision with other robots in red, and one can observe MRMP via parabolic gradually resolving the collisions and converging toward optimal solution that is safe and dynamically feasible.





\section{Conclusions}\label{conclusion}

In this paper, we present MRMP via parabolic relaxation that generates both optimal and dynamically feasible trajectories by planning in the coupled continuous joint-space of all robots. Our approach overcomes limitations that the state-of-the-art methods have which rely on de-coupling or local approximations, and is capable of finding optimal solutions to challenging motion planning problems. Through a multi-stage optimization approach that leverages low complexity, we attain computational tractability for problems involving a hundred robots in a highly dense environment. 
In the future work, we plan to extend our approach to nonlinear systems and further reduce computation run-time through parallelization and distributed computation.





\section*{ACKNOWLEDGMENT}
This work was supported by the Jet Propulsion Laboratory’s Research and Technology Development (R\&TD) program. Part of the research was carried out at the Jet Propulsion Laboratory, California Institute of Technology, under a contract with the National Aeronautics and Space Administration. \textcopyright 2021 California Institute of Technology. Government sponsorship acknowledged.

\bibliographystyle{IEEEtran}
\bibliography{References}

\end{document}